\documentclass[twocolumn]{article}
\usepackage{graphicx,amsmath,booktabs}
\usepackage{xcolor}
\usepackage[colorlinks=true, linkcolor=black, citecolor=black, urlcolor=magenta]{hyperref}
\usepackage{xcolor}

\usepackage{pgfplots, pgfplotstable}
\usepackage{subcaption}
\usepackage{siunitx}
\pgfplotsset{compat=1.18}
\usepackage{pgfplots}
\pgfplotsset{compat=1.18}
\usepackage{graphicx}
\usepackage{subcaption}
\usepackage{dblfloatfix}   % enables [b] for figure*/table*
\usepackage{microtype}     % nicer line breaks (optional)

\usepackage{booktabs,threeparttable}

\usepackage{comment} %\begin{comment} and \end{comment} to comment out
\usepackage{authblk}

\usepackage[
  backend=biber,
  style=numeric,      % numeric/Vancouver-style (common in med/CS)
  sorting=none,       % cite in order of appearance
  maxcitenames=2,     % “et al.” after 2 authors
  doi=true,isbn=false,url=false
]{biblatex}
\title{Fully Automated Deep Learning–Based Glenoid Bone Loss Measurement and Severity Stratification on 3D CT in Shoulder Instability}

\author[1]{Zhonghao Liu}
\author[2]{Hanxue Gu}
\author[3]{Qihang Li}
\author[6]{Michael Fox}
\author[6]{Jay M.~Levin}
\author[2,3,4,5]{Maciej A.~Mazurowski}
\author[6]{Brian C.~Lau}

\affil[1]{Department of Biomedical Engineering, Duke University%,
}
\affil[2]{Department of Electrical and Computer Engineering, Duke University%,
}
\affil[3]{Department of Biostatistics and Bioinformatics,
Duke University%,
}

\affil[4]{Department of Computer Science, Duke University%,
}

\affil[5]{Department of Radiology, Duke University%,
}

\affil[6]{Department of Orthopaedic Surgery,
Duke University Medical Center%,
}

\begin{document}
\maketitle

\begin{abstract}

\textbf{Purpose:} To develop and validate a fully automated, deep-learning pipeline for measuring glenoid bone loss on 3D CT scans using linear-based, en-face view, and best-circle method.

\noindent\textbf{Material and Methods:} Shoulder CT scans of 81 patients were retrospectively collected between January 2013 and March 2023. Our algorithm consists of three main stages: (1) Segmentation, where we developed a U-Net to automatically segment the glenoid and humerus; (2) anatomical landmark detection, where a second network predicts glenoid rim points; and (3) geometric fitting, where we applied a principal component analysis (PCA), projection, and circle fitting to compute the percentage of bone loss. The performance of the pipeline was evaluated using DSC for segmentation and MAE and ICC for bone-loss measurement; intermediate outputs (rim point sets and en-face view) were also assessed.

\begin{comment}
    using the Dice similarity coefficient (DSC) for segmentation, the Chamfer distance for glenoid rim point sets, the angular difference for en-face view of the glenoid, and the mean absolute error (MAE) and intraclass correlation coefficient (ICC) for the bone loss measurement.
\end{comment}

\noindent\textbf{Results:} Automated measurements showed strong agreement with consensus readings, exceeding surgeon-to-surgeon consistency (ICC 0.84 vs 0.78 for all patients; ICC 0.71 vs 0.63 for low bone loss; ICC 0.83 vs 0.21 for high bone loss; $P < .001$). For the classification task of assigning each patient to different bone loss severity subgroups, the pipeline’s sensitivity was 71.4\% for the low-severity group and 85.7\% for the high-severity group, with no instances of misclassifying low as high or vice versa.

\noindent\textbf{Conclusion:} A fully automated, deep learning–based pipeline for glenoid bone-loss measurement on CT scans can be a clinically reliable tool to assist clinicians with preoperative planning for shoulder instability. We are releasing our model and dataset at \url{https://github.com/Edenliu1/Auto-Glenoid-Measurement-DL-Pipeline}.

\end{abstract}

\noindent\textbf{Abbreviations}: ICC = intraclass correlation coefficient, DSC = dice similarity coefficient, MAE = mean absolute Error

\noindent\textbf{Summary:} A fully automated, deep learning–based pipeline can accurately predict glenoid bone loss measurements and classify bone loss severity, providing clinical decision support to assist clinicians with preoperative planning for shoulder instability.

\noindent\textbf{Key Points:} 

The pipeline showed good overall reliability relative to the physicians’ consensus and slightly outperformed the inter-reader (human--human) baseline (ICC 0.84 vs 0.78). The ICCs were 0.71 vs 0.63 for the low--bone-loss subgroup and 0.83 vs 0.21 for the high--bone-loss subgroup.

For classifying bone-loss severity, the model achieved a sensitivity of 71.4\% in the low-bone-loss subgroup and 85.7\% in the high--bone-loss subgroup, indicating robust performance in these two groups.

The algorithm’s strong classification performance in the low- and high-severity subgroups and its higher agreement with physician consensus than the inter-reader baseline suggest that the algorithm is clinically reliable for automating quantitative measurements, while human review remains important for intermediate-severity cases.

\section{Introduction}

The glenoid fossa is the shallow bony socket on the lateral scapula that articulates with the humeral head, allowing a wide range of arm motion. This mobility comes at the cost of stability\cite{Simonet1984}\cite{Boone2010}; the shoulder is therefore especially vulnerable to traumatic dislocation\cite{Owens2009}. Approximately 1.7\% of the general population has a traumatic shoulder instability, and the recurrence rate is substantial\cite{Simonet1984}\cite{Boone2010}\cite{Griffith2019}. In a clinical setting, accurate glenoid bone loss measurement is important for shoulder surgical planning, as an inaccurate measurement may lead to a completely different surgical management for the patient\cite{Paul2024}. To illustrate, smaller lesions with low bone loss may require conservative interventions, whereas larger defects may require surgical interventions such as laterjet transfers or allograft augmentations\cite{Thacher2023}.

There are many methods for measuring glenoid bone loss, 2D, 3D, linear-based, area-based, and volume-based, using CT or MRI\cite{Thacher2023}\cite{Rerko2013}\cite{eSouza2014}\cite{Huijsmans2011}. Here, our study focuses on 3D-reconstructed CT linear-based en-face views with a best-fit circle method, as it's the most popular method surgeons use for surgical planning\cite{Sugaya2003}\cite{Baudi2013pico}. The detailed procedure is as follows: (1) On an en-face view of a 3D bone rendered CT image, fit a circle to the glenoid 3 to 9 o’clock inferior contour. (2) Draw a line from the glenoid defect to the anterior margin of the circle (B). (3) Compute bone loss using: (B/A(diameter)) x 100\%. The most popular approaches for measuring glenoid bone loss include software-assisted (semi-automated) methods and fully manual measurements. Semi-automated tools typically standardize image preprocessing and segmentation but still rely on the clinician to select the en face view and fit the circle, and fully manual measurement is based on the doctor's best judgment. The weaknesses of both methods include extensive human labor and time, along with poor interrater variability (low agreement between doctors in en face view choice and bone-loss measurement)\cite{weaknessBois2012}\cite{weaknessKarpinski2024}\cite{weaknessLaunay2021}.

Many ideas have been proposed to solve these weaknesses; only a few are fully automated. Haimi’s method segmented the glenoid using a Hounsfield units (HU) threshold within an interval, applied PCA to reorient the glenoid slices to an en-face view, and fit a circle using a few predetermined clock positions on the glenoid segmentation \cite{Haimi2024Automated_2d}. The flaws of this method are that HU-based segmentation may inaccurately include pixels of arthritis that fall within the interval; PCA is not the optimal solution for finding the en-face view because it can be largely influenced by cupping bone loss, which is discussed later in the paper in Appendix~\ref{app:appendixC}; and the bone loss measurement was computed over a selected 2D slice of a CT scan, which many studies have found can underestimate the bone loss\cite{Huijsmans2011}. Zhao et al.’s method requires a healthy contralateral side for the bone loss computation, while many CT scans include only one side\cite{Zhao2023}.

We propose a fully automated, deep learning–based pipeline for 3D CT glenoid segmentation and measurement of glenoid bone loss using a linear-based, en-face view and the best-fit circle method. The glenoid bone is obtained using a deep learning model that is not subect to shoulder arthritis. A better approach is used to find the en face view, relying only on information from the intact side of the glenoid rim, which enables identification of the same en-face view of across different shapes of glenoid bone loss. Bone loss is measured on a 3D bone-rendered image rather than a 2D slice to avoid underestimation. A healthy contralateral side is not needed for this pipeline. Moreover, we assessed agreement using intraclass correlation coefficients (ICCs) to ensure that measurements to quantify inter-rater agreement across raters (algorithm vs. gold standard), which matters for understanding reliability of algorihtm for making surgical decisions. We also analyzed three bone-loss severity subgroups to approximate surgical decision management. We found that our pipeline for automating glenoid bone loss achieves higher agreement and correlation than the human baseline and demonstrates good reliability, especially in subgroups with low and high severity of bone loss. It is clinically reliable for assisting physicians in preoperative planning.

The purpose of this study is to develop and validate a fully automated, deep-learning pipeline for measuring glenoid bone loss on 3D CT scans using glenoid rim points, geometric fitting, and a principle component analysis, projection, and circle fitting to compute the percentage of bone loss. We hypothesize that automated measurements would demonstrate comparable performance to surgeon measurements.

\begin{figure*}[thp]
  \centering
  \includegraphics[width=0.8\linewidth]{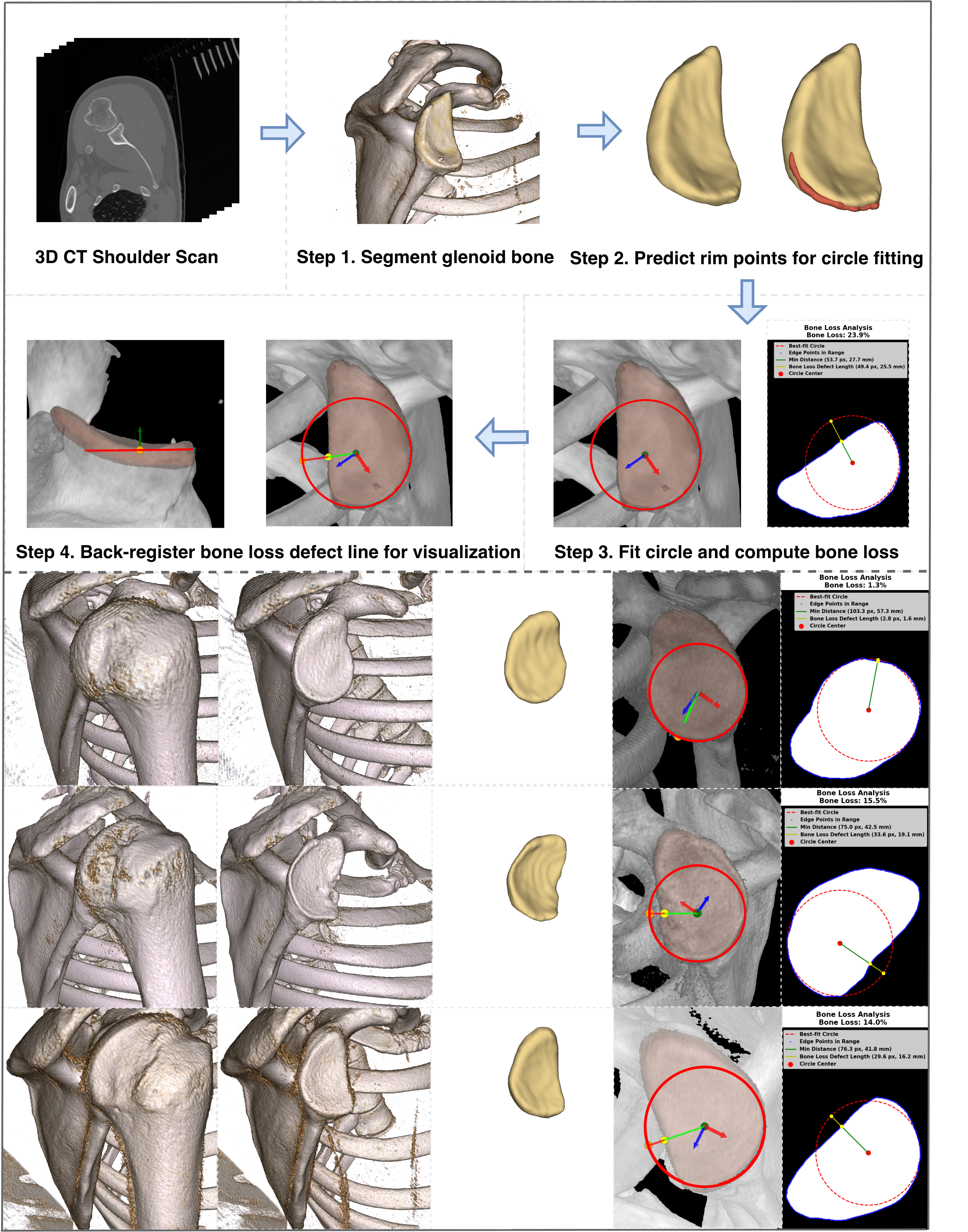}
  \caption[Automated glenoid bone-loss pipeline]{Pipeline for automated glenoid bone loss measurement. 
  (1) The segmentation model delineates the glenoid articular surface; 
  (2) \textsc{RimU-Net} reads the segmentation mask and predicts rim points; 
  (3) singular value decomposition (SVD) estimates the rim-plane normal, defining the en face view; 
  (4) the segmentation mask and rim points are projected onto the en-face plane; and 
  (5) a 2D circle is fitted to the rim points. 
  Bone loss is computed as $100\times B/A$, where $B$ is the glenoid-defect length and $A$ is the fitted-circle diameter.} Three more examples for demonstration.
  \label{fig:auto_glenoid}
\end{figure*}

% Make sure you have \usepackage{booktabs} in your preamble
\begin{table*}[ht]
\centering
\caption{Distribution of anterior glenoid bone-loss categories}
\label{tab:classdist}
\begin{tabular*}{\linewidth}{@{\extracolsep{\fill}} lccc}
\toprule
\textbf{Bone-loss category} & \textbf{Train (n = 60)} & \textbf{Test (n = 21)} & \textbf{Total (n = 81)} \\
\midrule
$<$\,13.5\,\%               & 26\,(43\%) & 7\,(33\%) & 33\,(41\%) \\
13.5--20\,\%                & 15\,(25\%) & 7\,(33\%) & 22\,(27\%) \\
$>$\,20\,\%                 & 19\,(32\%) & 7\,(33\%) & 26\,(32\%) \\
\midrule
\textbf{Total}              & 60\,(100\%) & 21\,(100\%) & 81\,(100\%) \\
\bottomrule
\end{tabular*}
\end{table*}

\section{Materials and Methods}

\subsection{Patient Data}
This study was conducted in accordance with ethics guidelines and received approval from the institutional review board (Pro10080765). Informed consent was waived due to the retrospective design of the study.

All CT scans were de-identified prior to analysis using a published DICOM metadata and burned-in text de-identification workflow \cite{Macdonald2024}

Our dataset of 3D shoulder CTs were retrospectively collected from patients between January 2013 and March 2023. The training dataset was sampled from this entire dataset shoulder CT dataset and was screened and labeled by an orthopedic surgeon (over 15 years of experience). The test dataset was sampled from a cohort within the same dataset but comprising healthy patients and patients who had undergone shoulder stabilization surgery, and it was screened and labeled by two experienced surgeons. Exclusion criteria included: (1) patients with posterior glenoid bone loss, (2) patients with a shoulder implant, and (3) patients with severe arthritis (measurement not feasible for physicians). Further dataset details are provided in Appendix~\ref{app:appendixA}.

Different degrees of glenoid bone loss require different preoperative planning. There is a general consensus that glenoid defects beyond roughly 20\% of the glenoid diameter should be managed with a bony augmentation procedure – most commonly the Latarjet coracoid transfer or a similar bone graft technique – rather than with labral repair alone\cite{Park2020}\cite{Keeling2023}. Therefore, we considered it important to test the algorithm in groups representing different levels of bone loss: $<13\%$, $13-20\%$, and $>20\%$. 
\begin{comment}
However, random sampling would heavily over-represent cases with $<13\%$ bone loss (majority class), causing a pronounced class imbalance, since only 1.7\% of the general population experiences traumatic shoulder instability\cite{Simonet1984}\cite{Boone2010}\cite{Griffith2019}. To mitigate this bias, an orthopedic surgeon randomly selected cases from patients who had undergone shoulder surgery and manually measured glenoid bone loss. To even further facilitate the data collection process, we trained the pipeline with around 20 training examples, with clinician verification, to identify additional patients with severe bone loss to complete our data set formation.
After excluding cases with posterior glenoid bone loss and evenly distributing them to meet criteria for three categories, we finalized the dataset with 98 shoulder CTs.
\end{comment}

We stratified the screened dataset and randomly divide it into a 20\% training data set and a 80\% test data set, resulting in 21 patient CTs in the test data set and 60 in the train data set, as shown in Table \ref{tab:classdist}. Demographic characteristics and scanner's information of the trainning and testing cohorts can be found in Table~\ref{tab:demographics_scanner}.

% Make sure you have \usepackage{booktabs} in your preamble
\begin{table*}[ht]
\centering
\caption{Distribution of demographics and scanner characteristics}
\label{tab:demographics_scanner}
\begin{tabular*}{\linewidth}{@{\extracolsep{\fill}} lccc}
\toprule
\textbf{Characteristic} & \textbf{Train (n = 60)} & \textbf{Test (n = 21)} & \textbf{Total (n = 81)} \\
\midrule
\textbf{Sex} & & & \\
Male                   & 41\,(68\%)  & 16\,(76\%)  & 57\,(70\%) \\
Female                 & 19\,(32\%)  & 5\,(24\%)   & 24\,(30\%) \\
\midrule
\textbf{Age (years)} & & & \\
Mean $\pm$ SD         & 44.3 $\pm$ 22.6 & 28.5 $\pm$ 13.1 & 40.2 $\pm$ 21.6 \\
Range                 & [14, 89]    & [15, 65]    & [14, 89] \\
Missing               & 2           & 1           & 3 \\
\midrule
\textbf{Weight (kg)} & & & \\
Mean $\pm$ SD         & 81.8 $\pm$ 17.1 & 83.7 $\pm$ 16.4 & 82.4 $\pm$ 16.8 \\
Range                 & [45.7, 133.0] & [59.0, 113.4] & [45.7, 133.0] \\
Missing               & 24          & 3           & 27 \\
\midrule
\textbf{Scanner model} & & & \\
SOMATOM Force              & 18\,(30\%) & 9\,(43\%)  & 27\,(33\%) \\
Discovery CT750 HD         & 15\,(25\%) & 10\,(48\%) & 25\,(31\%) \\
SOMATOM Definition Flash   & 9\,(15\%)  & 2\,(10\%)  & 11\,(14\%) \\
Revolution HD              & 10\,(17\%) & 0\,(0\%)   & 10\,(12\%) \\
SOMATOM Definition AS      & 4\,(7\%)   & 0\,(0\%)   & 4\,(5\%) \\
LightSpeed VCT             & 3\,(5\%)   & 0\,(0\%)   & 3\,(4\%) \\
Optima CT660               & 1\,(2\%)   & 0\,(0\%)   & 1\,(1\%) \\
\midrule
\textbf{Total patients}    & 60\,(100\%) & 21\,(100\%) & 81\,(100\%) \\
\bottomrule
\end{tabular*}
\end{table*}

\subsection{Automated Glenoid Bone Loss Measurement Pipeline}
\begin{figure}[ht]
  \centering
  \includegraphics[width=\linewidth]{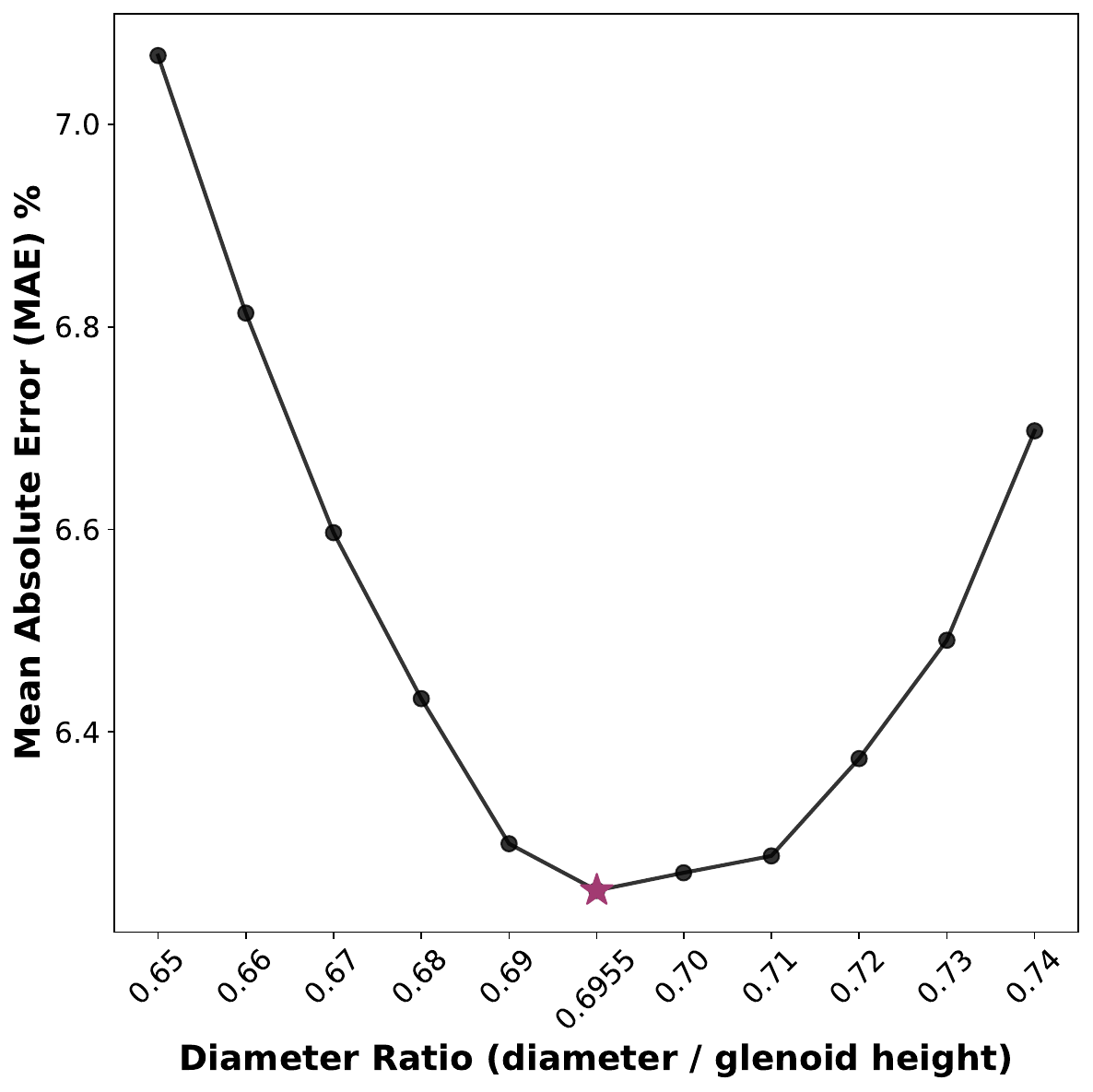}  
    \caption{Optimization of Diameter Ratio for Glenoid Bone Loss Prediction on Training Dataset.
    The X-axis represents the diameter ratio (fitted circle diameter divided by glenoid height) tested across 11 values from 0.65 to 0.75; the Y-axis represents the mean absolute error (MAE, \%) in bone loss prediction.
    The purple star ($\star$) marks the optimal diameter ratio of 0.6955, which achieved MAE = 6.24\%.
    For comparison, the unconstrained fitting method achieved MAE = 8.81\%.}
    \label{fig:best_diameter}
\end{figure}

We introduce a fully automated, end-to-end deep learning-based pipeline for segmenting the glenoid articular surface and measuring bone loss on 3D reconstructed CT using linear-based, en-face view, and best-fit circle method. The pipeline follows the same procedure applied by
the clinician, as described in the introduction. Below is a brief pipeline description: 1. The trained deep learning segmentation model reads the shoulder CT scan and outputs the left/right glenoid and humerus bone loss labels. 2. The glenoid bone label is flipped along the depth dimension if detected as a Right glenoid. 3. The trained landmark neural network model (RimU-Net) reads the preprocessed binary segmentation map from the segmentation model and predicts the rim point sets. The predicted rim points are then back-registered to the original segmentation map 4. Given 3-D rim-skeleton points, we estimate a best-fit plane via SVD/PCA on the point cloud, taking the direction normal to the top two largest variance directions as the en-face view. 5. Glenoid segmentation and rim points are projected onto this plane. A 2-D geometric circle is fit by minimizing the radial least-squares objective $J(c,r)=\sum_i (\lVert p_i-c\rVert_2-r)^2$ using BFGS (c = center of circle coordinate; r = radius of the circle; p = rim point sets). Based on the clinical suggestion of the diameter of the fitted circle usually being 2/3 of the glenoid height, we did an optimization test using a grid search around the hyperparameter of 2/3 on the train set. Here we explicitly set the radius to be 0.6955/2 based on the analysis (Fig.~\ref{fig:best_diameter}). 6. Bone loss is measured by dividing the glenoid defect (B) by the diameter of the fitted circle (A): $(B/A) \times 100\%$. Following Hami's method, determining B, many radial lines are drawn from the glenoid defect to the anterior margin of the circle, and the longest line represents the length of the glenoid defect \cite{Haimi2024Automated_2d}. A general workflow of this fully automated pipeline is outlined in the flowchart provided in Figure \ref{fig:auto_glenoid}.

\subsection{Data Annotation}
\textbf{Segmentation annotation}: The annotation of the scapulae and humerus bones for the entire dataset was completed under the guidance of an orthopedic surgeon with more than 10 years of experience in sports medicine. We segmented the thin layer at the top of the glenoid fossa, which experts would consider a part to measure bone loss in the glenoid after finding the en-face view. All segmentation annotations were refined and approved by one surgeon for the training dataset and by two surgeons for the test dataset.

\noindent\textbf{Landmark annotation}: A variable number of landmarks are selected on the posterior–inferior rim surface of the annotated binary glenoid segmentation mask in 3D Slicer, and these landmarks are used to train the circle-fitting model. The rim curve used for circle fitting is shorter in patients with substantial glenoid bone loss and longer in patients with minimal bone loss. Therefore, we train a model that can accommodate varying degrees of bone loss by outputting different sets of rim points. Researchers and surgeons are provided with instructions for selecting landmarks that best represent the intact posterior–inferior glenoid rim.

\begin{enumerate}
  \item Rotate the glenoid surface mask until the en-face view is perpendicular to the labeler’s viewing plane \cite{Sugaya2003}.
  \item Based on the labeler’s experience, select a landmark at the approximate 3 o’clock position that corresponds to the intact posterior–inferior rim \cite{Sugaya2003}.
  \item Following the outermost rim, continue selecting landmarks until the curve reaches the inferior aspect of the defect \cite{Chen2022}.
\end{enumerate}

Similar to the segmentation annotations, landmarks were manually labeled and then refined and approved by an experienced surgeon. One surgeon refined and approved the training dataset, whereas two surgeons refined and approved the test dataset.

\subsection{Model's Architecture}
\textbf{Segmentation backbone (nnU-Net / TotalSegmentator)}. We only have 61 volumes in training data, which is considered a low number of images for deep learning training. To address this data scarcity, we conducted a two-stage fine-tuning of the public TotalSegmentator model, built on nnU-Net \cite{Isensee2020} \cite{Wasserthal2023}. Transfer learning is a highly data-efficient approach for training models in small subsets of data \cite{Pan2010}, and TotalSegmentator is a multi-label segmentation model trained on 104 anatomic structures across 1204 CT examinations, encompassing the scapula and humerus bones. We believe that fine-tuning Totalsegmentator can transfer its rich anatomical knowledge of these bones in the CT domain to our task of segmenting four target labels: left/right glenoids and left/right humerus bones (Appendix~\ref{app:appendixA}).

\noindent\textbf{RimU-Net: Heatmap regression model for circle fitting}.

RimU-Net is a U-Net based heatmap regression network designed for thin-curve localization. Given the glenoid surface segmentation, it predicts a heatmap/point set corresponding to the intact posterior–inferior rim, which is then used to estimate the best-fit circle. Architecture and training details are provided in Appendix~\ref {app:appendixB}.

Manual measurement of glenoid bone loss in 3D CT scans by doctors in general requires (1) finding an en-face view, (2) projecting the 3D image in that direction, and (3) drawing a best-fit circle to the outermost rim/contour \cite{Sugaya2003}. This pipeline would require estimating two quantities: the direction vector at which a doctor would think makes the glenoid surface normal to their view (en-face view), and the rim curve at the intact glenoid contour (excluding bone loss areas), which they refer to for circle fitting. Inspired by the Constellation Technique (CST) \cite{Chen2022}, which labels the glenoid rim regardless of en-face orientation, we introduce a novel approach to train a deep learning model to estimate the best-fit circle on the glenoid contour, solving both tasks above with a single model using a small dataset. Rim data for training is selected first, reorienting the en-face view, and then, on that plane, select the outermost rim, encompassing information about both rim and the en-face view coordinates. Therefore, RimNet can skip the en-face view localizing and directly predict the outermost rim contour that the doctors use for circle fitting in 3D space. The en-face view direction vector can then be calculated with SVD on the curve points.

\subsection{Evaluation}

To evaluate the proposed deep-learning pipeline for glenoid bone measurement, we designed experiments focusing on (i) segmentation accuracy, (ii) accuracy of en-face view plane computation, (iii) accuracy of the fitted curve along the intact posterior-inferior glenoid contour on the en-face view, and (iv) accuracy of the final glenoid bone-loss measurement. We evaluated pipeline predictions on the test set across three bone-loss severity levels: $< 13.5\%$, $13.5\% < \text{loss} < 20\%$, and $> 20\%$.

\section{Results}

% Overall summary table (updated values)
\begin{table*}[t]
\centering
\begin{threeparttable}
\caption{Interobserver reliability between the automated pipeline and manual measurements of glenoid bone loss\textsuperscript{a}}
\label{tab:icc_summary}
\begin{tabular*}{\textwidth}{@{\extracolsep{\fill}} llll l}
\toprule
Comparison & Category & ICC (95\% CI) & Pearson $r$ & $n$ \\
\midrule
Doctor 1 vs Doctor 2               & Human Baseline         & 0.7822 (0.5292, 0.9075) & 0.7940 & 21 \\
Algorithm vs Doctor 1              & Algorithm vs Doctor    & 0.7842 (0.5329, 0.9084) & 0.7949 & 21 \\
Algorithm vs Doctor 2              & Algorithm vs Doctor    & 0.8036 (0.5695, 0.9171) & 0.8088 & 21 \\
\textbf{Algorithm vs Consensus}             & \textbf{Algorithm vs Consensus} & \textbf{0.8382 (0.6370, 0.9325)} & \textbf{0.8460} & \textbf{21}\\
\bottomrule
\end{tabular*}
\begin{tablenotes}[flushleft]
\footnotesize
\item \textsuperscript{a}Values are ICC(A,1) with 95\% confidence intervals. Pearson $r$ is the Pearson correlation coefficient. Consensus is the mean of doctors' measurements. $n$ indicates the number of paired measurements.
\end{tablenotes}
\end{threeparttable}
\end{table*}

% Subgroup table (no Pearson; updated values; four-decimal precision; parentheses for CI)
\begin{table*}[ht]
\centering
\begin{threeparttable}
\caption{ICC(A,1) by bone loss severity subgroup\textsuperscript{b}}
\label{tab:icc_subgroups}
\begin{tabular*}{\textwidth}{@{\extracolsep{\fill}} l l l l l}
\toprule
Group & Comparison & Category & ICC (95\% CI) & $n$ \\
\midrule
\multicolumn{5}{l}{\textbf{Low Bone Loss} (\textless{}13.5\%)} \\
\addlinespace[2pt]
& Doctor 1 vs Doctor 2          & Human Baseline         & 0.6259 (-0.2405, 0.9372) & 7 \\
& Algorithm vs Doctor 1         & Algorithm vs Doctor    & 0.6821 (-0.1460, 0.9481) & 7 \\
& Algorithm vs Doctor 2         & Algorithm vs Doctor    & 0.6638 (-0.1786, 0.9446) & 7 \\
& \textbf{Algorithm vs Consensus}        & \textbf{Algorithm vs Consensus} & \textbf{0.7091 (-0.0944, 0.9532)} & \textbf{7} \\
\addlinespace[4pt]
\multicolumn{5}{l}{\textbf{Moderate Bone Loss} (13.5--20\%) } \\
\addlinespace[2pt]
& Doctor 1 vs Doctor 2          & Human Baseline         & -0.1439 (-0.8093, 0.6832) & 7 \\
& Algorithm vs Doctor 1         & Algorithm vs Doctor    & -0.0664 (-0.7804, 0.7228) & 7 \\
& Algorithm vs Doctor 2         & Algorithm vs Doctor    & -0.1190 (-0.8003, 0.6965) & 7 \\
& \textbf{Algorithm vs Consensus}        & \textbf{Algorithm vs Consensus} & \textbf{-0.0917 (-0.7902, 0.7104)} & \textbf{7} \\
\addlinespace[4pt]
\multicolumn{5}{l}{\textbf{High Bone Loss} (\textgreater{}20\%) } \\
\addlinespace[2pt]
& Doctor 1 vs Doctor 2          & Human Baseline         & 0.2171 (-0.6407, 0.8338) & 7 \\
& Algorithm vs Doctor 1         & Algorithm vs Doctor    & 0.5190 (-0.3842, 0.9146) & 7 \\
& Algorithm vs Doctor 2         & Algorithm vs Doctor    & 0.7619 (0.0208, 0.9626)  & 7 \\
& \textbf{Algorithm vs Consensus}        & \textbf{Algorithm vs Consensus} & \textbf{0.8326 (0.2132, 0.9746)}  & \textbf{7} \\
\bottomrule
\end{tabular*}
\begin{tablenotes}[flushleft]
\footnotesize
\item \textsuperscript{b}Values are ICC(A,1) with 95\% confidence intervals. Consensus is the mean of doctors' measurements. $n$ indicates the number of paired measurements in each subgroup.
\end{tablenotes}
\end{threeparttable}
\end{table*}

\subsection{Experiment 1: Evaluation of Glenoid Articular Surface Segmentation}

%0.8746 ± 0.0437 before   0.9866 \pm 0.0059 for humerus

The segmentation predicted by the model was compared with the manual labels from two clinicians. The average Dice similarity coefficient was $0.875 \pm 0.044$ for the glenoid (Q1--Q3: $0.847$--$0.896$; range: $0.762$--$0.930$; $n=21$). 
For the humerus, it was $0.987 \pm 0.006$ (Q1--Q3: $0.981$--$0.992$; range: $0.972$--$0.994$; $n=21$).

\subsection{Experiment 2: Evaluation of Predicted Rim on the Intact Posterior Contour of the Glenoid} 
We evaluated the accuracy of the rim points predicted by RimU-Net using the bidirectional Chamfer distance between the predicted and ground-truth glenoid rim skeletons (in mm). The Chamfer distance estimates how similar the predicted rim point set is to the set annotated by the clinician on the glenoid surface. The mean nearest-neighbor distance was $1.37 \pm 0.58 \,\text{mm}$, with most cases having distances $< 1.17 \,\text{mm}$.

\subsection{Experiment 3: Assessment of En-face Direction Vector} 

\begin{figure}[htp]
  \centering
  \includegraphics[width=\linewidth]
  {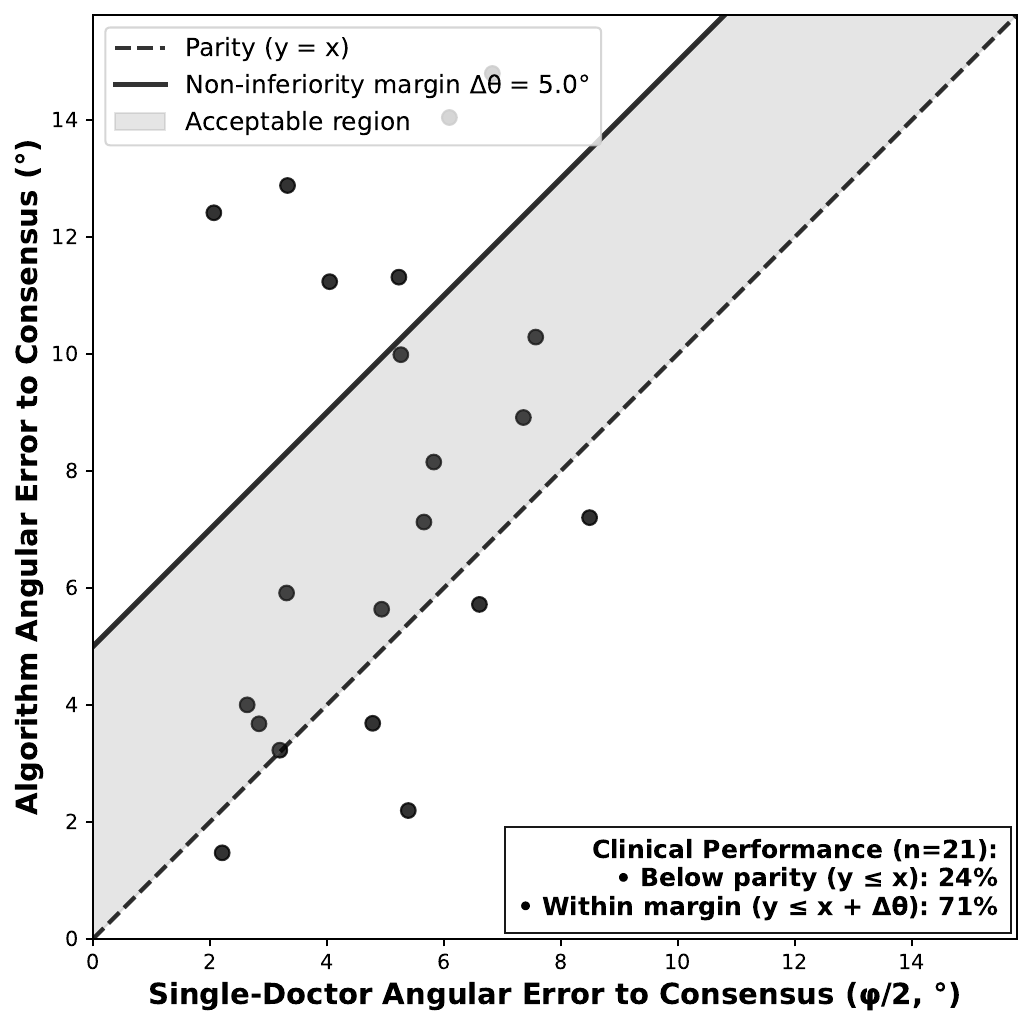}  
    \caption{Scatter plot comparing the algorithm-predicted en-face direction to the doctor baseline. Each point is one case.
    The X-axis is the error of one doctor's measurement to the consensus(two doctors' mean); the Y-axis is the predicted angular error to the same consensus.
    The dashed diagonal ($y{=}x$) marks parity with a typical single doctor; points below this line have lower error than a single doctor.
    The dotted line ($y{=}x{+}5^\circ$) indicates the pre-specified non-inferiority margin; points in the gray region at or below this line are \emph{within the allowable tolerance}.}
    \label{fig:direction_scatter}
\end{figure}

We evaluated the en-face view of the glenoid computed by the algorithm against manual selections from two clinicians. The mean of a doctor's angular error to the consensus on en face view is $4.94^\circ \pm 1.86^\circ$ (95\% CI: $4.09^\circ$--$5.78^\circ$), while the algorithm's angular error to consensus is $7.85^\circ \pm 3.99^\circ$ (95\% CI: $6.03^\circ$--$9.67^\circ$). While the mean error of algorithm-consensus pair is higher, of 21 observation analyzed, 24\% demonstrated algorithm performance below parity ($y \leq x$), and 71\% fell within the clinically acceptable non-inferiority margin of $5^\circ$ ($y \leq x + \Delta\theta$)(Fig~\ref{fig:direction_scatter}).

\subsection{Experiment 4: Glenoid Bone Loss}

We evaluated the accuracy and reliability of the fully automatic end-to-end pipeline. The algorithm demonstrated a mean absolute error of \(4.28 \pm 3.38\%\) (median \(= 3.79\%\)) when compared with the consensus ground truth (the mean of two expert assessments). Using Pearson’s correlation coefficient, we assessed the correlation between the pipeline’s predicted bone loss and the clinician’s measurements on the test set. Correlation between the algorithm and the consensus was \(r=0.846\) with \(p<0.001\), compared with the inter-rater correlation \(r=0.794,\, p<0.001\), showing an stronger linear correlation 

Table~\ref {tab:icc_summary} presents two-way mixed-effects, absolute-agreement, single-measure intraclass correlation coefficient ICC(A,1)(Table~\ref{tab:icc_summary}). Algorithm agreement with each clinicians was similar or slightly higher than the doctor-to-doctor baseline. Moreover, the algorithm achieved good reliability overall against consensus, slightly higher than the human-human baseline (ICC 0.84 vs 0.78) with an absolute increase of 0.06.

Intraclass correlation coefficient was also estimated on three bone loss severity subgroups (Table~\ref {tab:icc_subgroups}). Algorithm’s agreement was higher than the human–human baseline for all subgroups except for the moderate bone loss subgroup. In low bone loss group algorithm against concensus has ICC higher than human baseline (ICC 0.71 vs 0.63). In moderate bone loss group agreement across all comparions are near 0, which shows either no correlation between raters or that the variance between patients are too small. Importantly, in high bone loss group human baseline showed a poor reliability while algorithm-clinician showed a much higher agreement, especially that algorithm-consensus showed a good reliability (ICC 0.21 vs 0.83), an absolute increase of 0.62 (62 percentage points; 383.5\% relative).

\begin{figure*}[tp]
\centering

\begin{subfigure}{0.47\linewidth}
    \centering
    \includegraphics[width=\linewidth]{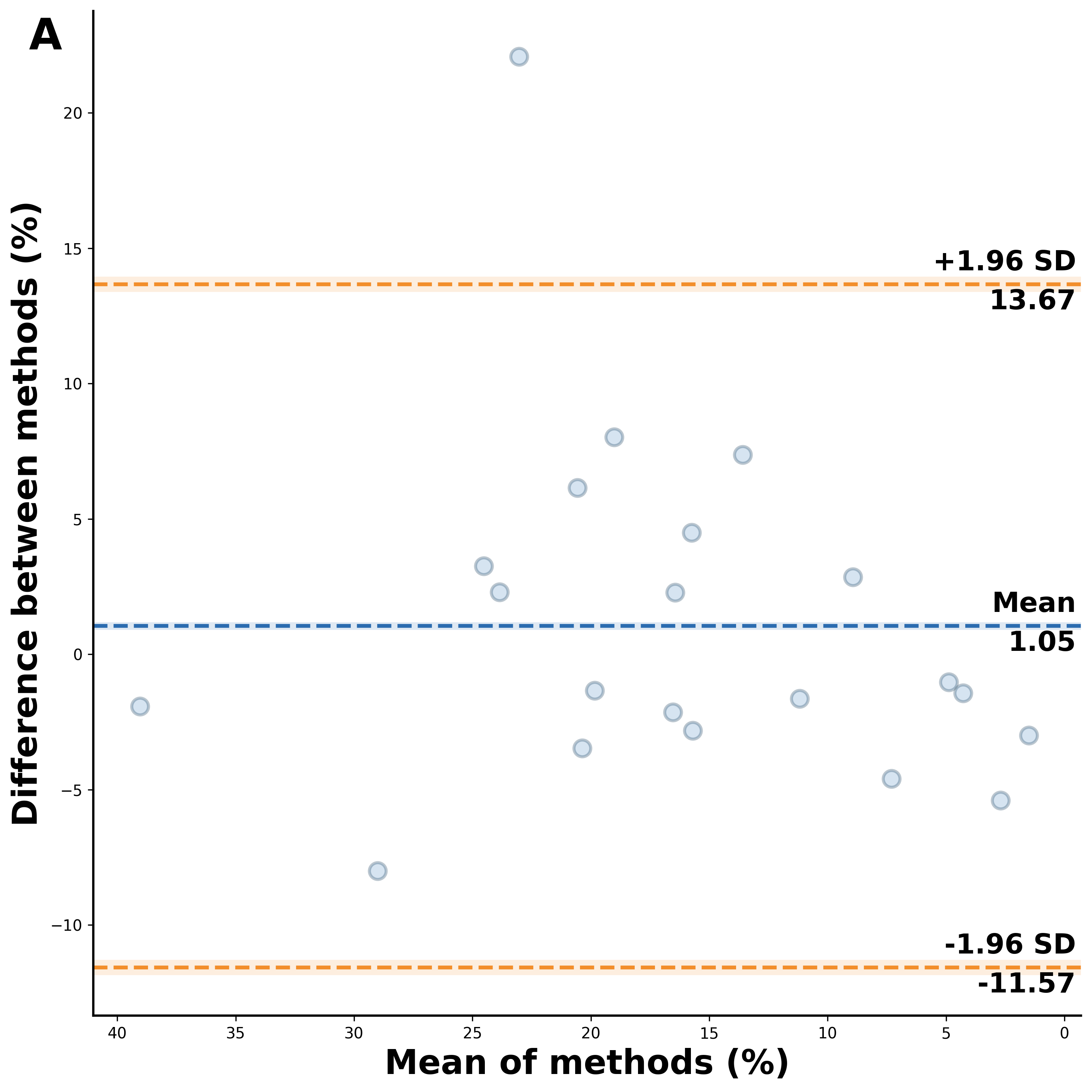}
    \caption{Doctor vs. doctor (n=21)}
    \label{fig:bland_altman_human}
\end{subfigure}\hfill
\begin{subfigure}{0.47\linewidth}
    \centering
    \includegraphics[width=\linewidth]{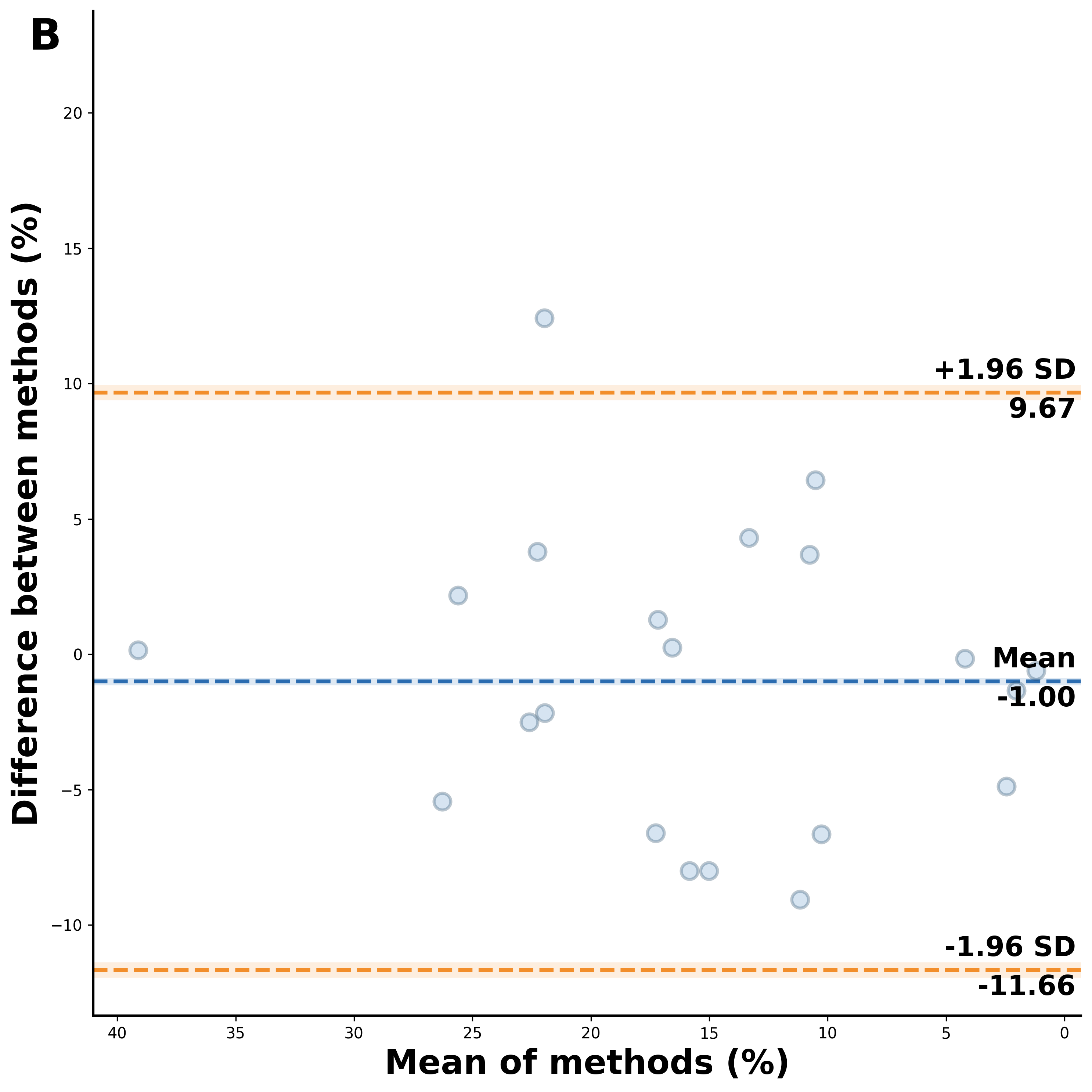}
    \caption{Algorithm vs. consensus (n=21)}
    \label{fig:bland_altman_algorithm}
\end{subfigure}

\caption{Bland--Altman plots showing agreement between two doctors alongside agreement between the algorithm and consensus measurements.}
\label{fig:bland_altman_combined}
\end{figure*}

Bland--Altman agreement for glenoid bone loss from the algorithm showed minimal systematic bias versus the clinician consensus (bias: $-0.997\%$). Relative to the human-to-human baseline, the algorithm exhibited similar bias and a narrower spread of differences (LoA width 21.33 \% vs 25.24 \%; absolute difference 3.91 \%). The algorithm’s SD of differences was correspondingly lower (5.44 vs 6.44 \%)(Fig~\ref{fig:bland_altman_combined}).

\begin{figure}[ht]
  \centering
  \includegraphics[width=\linewidth]
  {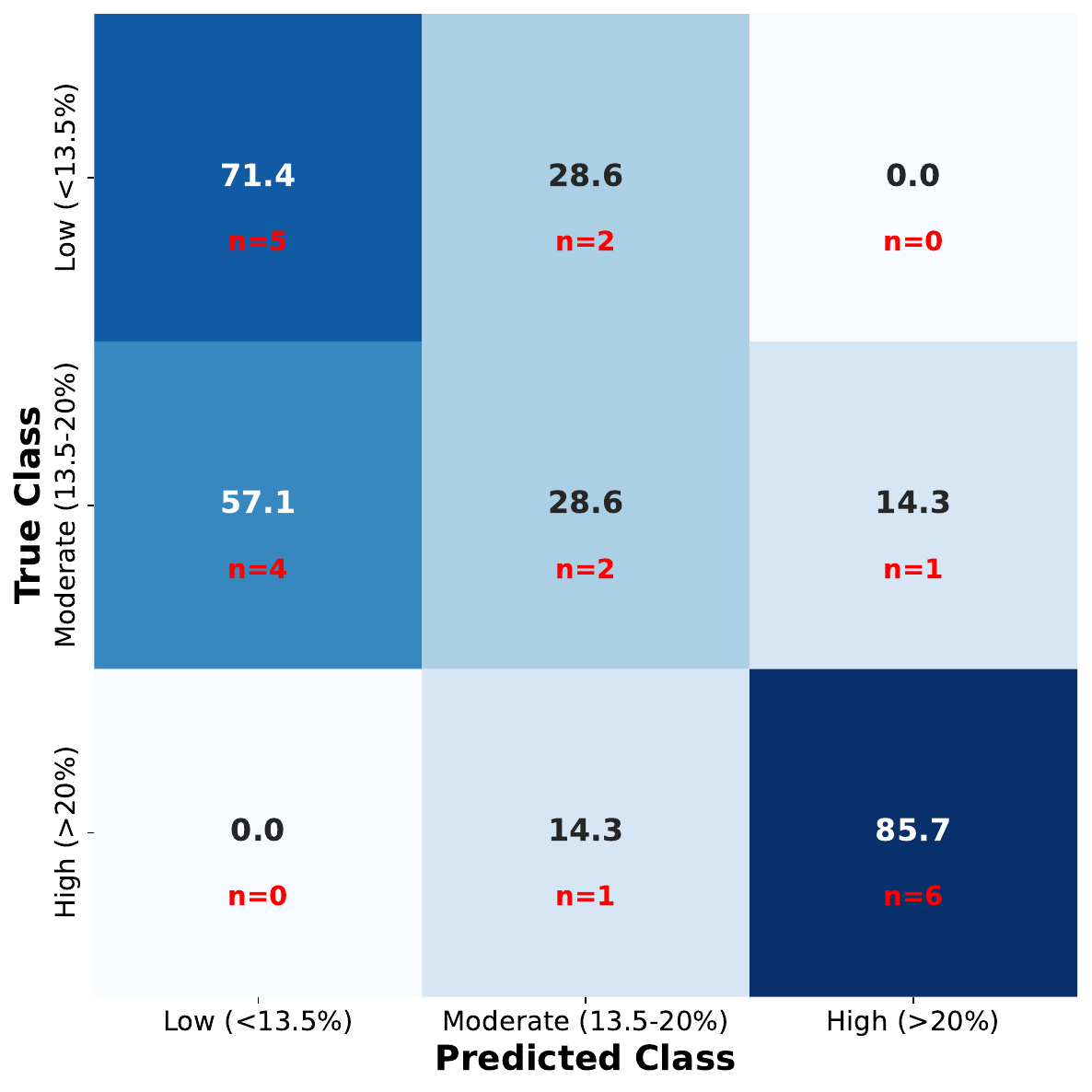}  
    \caption{Confusion matrix for the algorithm-predicted glenoid bone loss measurement against the doctor's consensus. The predicted class axis represents the algorithm's classification, and the true class represents the doctor's consensus. Accuracy is centered in each box, and the number of samples is categorized below.}
    \label{fig:confusion matrix}
\end{figure}

A confusion matrix was used to evaluate the ability of the pipeline to classify the severity of bone loss against the consensus' of the physicians (Fig.~\ref{fig:confusion matrix}). The Per-class recall (sensitivity) was 0.714 (5/7) for Low, 0.286 (2/7) for Moderate, and 0.857 (6/7) for High. Misclassifications were predominantly between adjacent categories (Low$\leftrightarrow$Moderate: 6 cases; Moderate$\leftrightarrow$High: 2 cases), with no Low$\leftrightarrow$High errors. For the clinically important High $(>20\%)$ class, recall was 0.857; that is, 6 of 7 High cases were correctly identified and the remaining case was labeled Moderate. The Moderate class had the lowest recall among all groups.

\section{Discussion}

The most important finding in this study is that (1) a fully automated deep learning-based algorithm pipeline can be used to reliabily automate the measurement of glenoid bone loss on 3D CT scans. (2) Our algorithm and pipeline can be a clinically reliable tool in assisting pre-operative planning in shoulder instability, especially in screening high bone loss ($>20\%$) as a classification model. (3). The segmentation model in the pipeline is also clinically reliable in subtracting the humerus bone and preparing the glenoid bone segmentation for glenoid bone-related assessment. Additionally, the current automated pipeline can reliably identify true en-face view.

In our analysis of bone loss, our automated pipeline demonstrated higher agreement and correlation with fellowship-trained orthopaedic surgeon measurements compared to the inter-rater baseline between surgeons. It is much better than the human baseline in the high bone loss group (ICC 0.83 vs. 0.22) in reliability. It shows minimal systematic bias and a smaller spread of differences than the human baseline, achieving 71.4 and 85.7 in classifying low and high bone loss groups, respectively, and without any instances of misclassifying low as high or vice versa. The correlation and classification in the moderate bone loss group are weak, but are consistent with the human baseline. All of which suggests that our deep learning-based algorithm can reliably automate the measurement of glenoid bone loss in CT scans. Our pipeline is clinically reliable for quantitative measurements and effectively identifies cases of high bone loss, although we recommend human review for cases within the 13.5\% to 20\% "gray zone." It is important to note that, in the gray zone, clinical surgeon decision-making remains gray as well. The current algorithm can alert surgeons that bone loss is within this zone and requires greater scrutiny, as would be the case with manual measurements. In areas of large bone loss, the algorithm accurately measures bone loss, which can allow surgeons to reliably plan bony augmentation techniques.

The humerus bone segmentation dice score of $0.9866 \pm 0.0059$ demonstrated that our segmentation model can be an excellent tool for humerus bone subtraction and, therefore, a great tool to help visualize the glenoid bone. As shown in Figure \ref{fig:auto_glenoid}, the humerus bone is completely removed, leaving no residual bone fragment and a complete view of the glenoid bone. It is important to note that current clinical CT 3D protocols require manual segmentation to remove humerus and create 3D rendered glenoid/scapula images. Based on the reliable bone loss measurements demonstrated by this pipeline and its overall clinical reliability for assessing glenoid bone loss, we believe the segmentation model, which is an essential component of the pipeline, is also reliable for supporting clinical decision making by providing glenoid visualization. This segmentation model may serve as a useful tool for preoperative planning in patients with shoulder instability.

\subsection{Limitations and future work}

The test dataset has small severity subgroups (n = 7/7/7), limiting statistical precision and generalizability. Consequently, estimates of agreement (e.g., Bland-Altman bias and limits of agreement) and reliability (ICC) have wide confidence intervals. We also did not validate the pipeline on the external dataset. Furthermore, the study only involved glenoid bones with anterior bone loss, but glenoid bone loss can also occur posteriorly. Magnetic resonance imaging is also commonly used to measure glenoid bone loss and has the advantage of being a noninvasive imaging modality. A future study can use an MRI image to automate the anterior and posterior measurement of glenoid bone loss and evaluate it on a larger test dataset and an external dataset.

\section{CONCLUSION}

To our knowledge, this is the first work to automate the measurement of glenoid bone loss on 3D reconstructed CT using linear-based, en-face view, and best-fit circle method. It segments the glenoid and subtracts the humerus bone accurately. This pipeline demonstrates good agreement and reliability with human annotation, and is even better than a human-to-human baseline. It is clinically reliable in automating quantitative measurements, especially in low($< 13.5\%$) and high bone loss($>20\%$) severity groups for the detection, while human review is required for cases between. In summary, it may be a robust tool in assisting doctors in pre-operative surgical planning for shoulder instability.

%\section*{Reference}
%\begin{doublespace}              % turn on double-spacing
\printbibliography

\appendix
\section{Appendix A}
\addcontentsline{toc}{section}{Appendix A}
\label{app:appendixA}

\begin{figure}[ht]
    \centering
    \includegraphics[width=\linewidth]{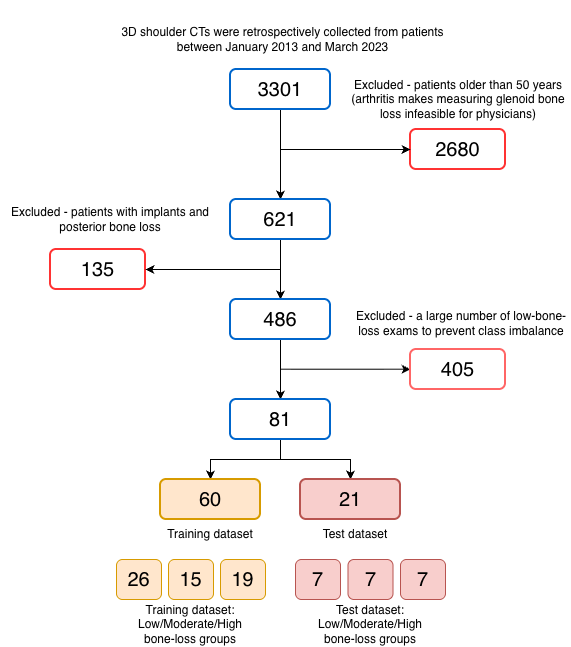}  
    \caption{Flowchart showing the datasets used to train and test the automated pipeline for measuring glenoid bone loss.}
    \label{fig:PCA_bad}
\end{figure}

\subsection{Ground-Truth Generation}
The ground truth for the segmentation model does not require any pre-processing, as the foreground binary mask can be used directly for training. In contrast, the ground truth for the rim model requires pre-processing, as there isn't a fixed number and location of landmarks that can represent the curve for circle fitting across different level of bone loss. To enable the model to learn from the selected landmarks and produce outputs that can be readily postprocessed for circle fitting and bone-loss computation, we designed the following steps: 1. For consistency, the landmarks for each case were resampled to 30 points in 3D Slicer. 2. Using the Markups to Model module in 3D Slicer, the landmark markup was converted to a tube with a 1.00-mm radius using a cardinal spline curve type. 3. The rim tube was then skeletonized into a thin curve with a thickness of 1 pixel. An isotropic 3D Gaussian kernel\(\sigma = 1\,\mathrm{mm}\) was applied to the skeletonized curve to generate a heatmap with values ranging from 0 to 1.

Additional pre-processing is used to inject an inductive bias and reduce data efficiency, leveraging the intrinsic geometry of the glenoid fossa. We believe that this can accelerate learning with limited data. To begin with, we applied PCA in each plate-shaped glenoid mask and rotated the volume so that the glenoid articular surface faces the z-axis, reducing the depth of the mask. With this, we can reduce the dimension of the glenoid binary mask from \(112^{3}\) voxels to \(48 \times 112 \times 112\). Second, all right shoulders are flipped across the sagittal plane so that they appears as a left shoulders, making it mirror invariance. These steps standardize laterality and reduce dimensionality, allowing the network to learn from a smaller, more simpler feature space.

\subsection{Two stages of fine-tuning}
The first stage of fine-tuning involves fully utilizing TotalSegmentator's large CT dataset and learning the coarse bony context of the scapula and humerus, which can later be refined on the substructures of the glenoid bone. We filtered out CTs that do not include the scapula label for the training dataset, leaving 849 training data. We trained the 3D full-resolution \texttt{nnU-Net} from TotalSegmentator (CT) using the trainer \texttt{nnUNetTrainerNoMirroring}, following no mirroring specification in its original training plan. Initialization used the public TotalSegmentator checkpoint \texttt{Dataset294\_TotalSegmentator\_part4\_muscles}, fold~0 (final checkpoint). Auto-planning selected 1.5~mm isotropic spacing, a \(128^3\)-voxel patch size, and a batch size of~2. We set the base learning rate at \(10^{-4}\) for fine-tuning; all other hyperparameters, augmentations, and schedules followed \texttt{nnU-Net} defaults.

In the second stage, we further fine-tuned the model on our sub-dataset to capture finer substructures, as described on the glenoid articular surface\cite{Chlebus2018}. We collected a smaller dataset under the guidance of an orthopedic surgeon, comprising 98 CT scans and their corresponding left/ right glenoid articular surfaces and humerus bone loss, as shown in Table \ref{tab:classdist}. The training plan used to fine-tune the first-stage model is identical to the one used in the first stage. We used the same loss function for both stages, L = LCE + LDICE, which combines entropy loss and the dice loss. We will release the exact plan/config JSON, trained weights, and scripts for full reproducibility.

\section{Appendix B}
\addcontentsline{toc}{section}{Appendix B}
\label{app:appendixB}

The RimU-Net is adapted from the structure of a vanilla U-net tailored for the heat-map regression thin-curve task. In addition to a basic encoder-decoder structure, we used strided convolution $(2,2,1)$ to replace the max-pooling for downsampling in the first two stages to preserve the details of the plane for our anistropic medical images \cite{Isensee2020}. Moreover, for each convolution, we use Instance Normalization for medical images in small batches \cite{Verweij2020} and LeakyReLU activations with $\alpha=0.01$ to preserve all gradients flowing to the prediction of our sparse target output \cite{Maas2013}. We used adaptive wing loss for our landmark localization task, as a greater penalty is needed for medium to small errors.  \cite{Feng2018Wing}\cite{Wang2019AdaptiveWing}.
\begin{equation}
\begin{aligned}
\operatorname{AWing}(y,\hat{y}) &= {} \\[3pt]
&\hspace{-2em}\begin{cases}
\omega\, \ln\!\left(
  1 + \left|\dfrac{y-\hat{y}}{\epsilon}\right|^{\alpha - y}
\right), & \text{if } |y-\hat{y}| < \theta,\\[6pt]
A\,|y-\hat{y}| - C, & \text{otherwise.}
\end{cases}
\end{aligned}
\end{equation}

where $A=\omega(1/(1+(\theta/\epsilon)^{\alpha-y}))(\alpha-y)((\theta/\epsilon)^{\alpha-y-1})(1/\epsilon)$ and $C=(\theta A-\omega\ln(1+(\theta/\epsilon)^{\alpha-y}))$.
We heuristically use parameters based on the Adaptive Wing Loss paper: \(\omega=16\), \(\epsilon=1\), \(\theta=0.5\), \(\alpha=2.1\) \cite{Wang2019AdaptiveWing}. We used Adam to optimize training and ReduceLROnPlateau built in PyTorch to dynamically adjust the learning rate \cite{Kingma2015Adam}. We use the symmetric Chamfer distance between the predicted and reference set of skeleton points as a metric to evaluate the performance of the model $\mathrm{CD}(A,B)=\frac{1}{|A|}\sum_{a\in A}\min_{b\in B}\lVert a-b\rVert_2+\frac{1}{|B|}\sum_{b\in B}\min_{a\in A}\lVert b-a\rVert_2$ \cite{chamberclassic}\cite{chambersymmetric}. The predicted heat maps are binarized at 0.3 (threshold chosen in the validation set), forming a tubular rim volume that is then skeletonized to a 1-voxel-wide curve using the 3-D thinning method of Lee et al.\cite{Lee1994SkeletonModels}. Chamfer distance is computed in millimeters via the voxel spacing.

\section{Appendix C}
\addcontentsline{toc}{section}{Appendix C}
\label{app:appendixC}

\begin{figure}[tph]
    \centering
    \includegraphics[width=\linewidth]{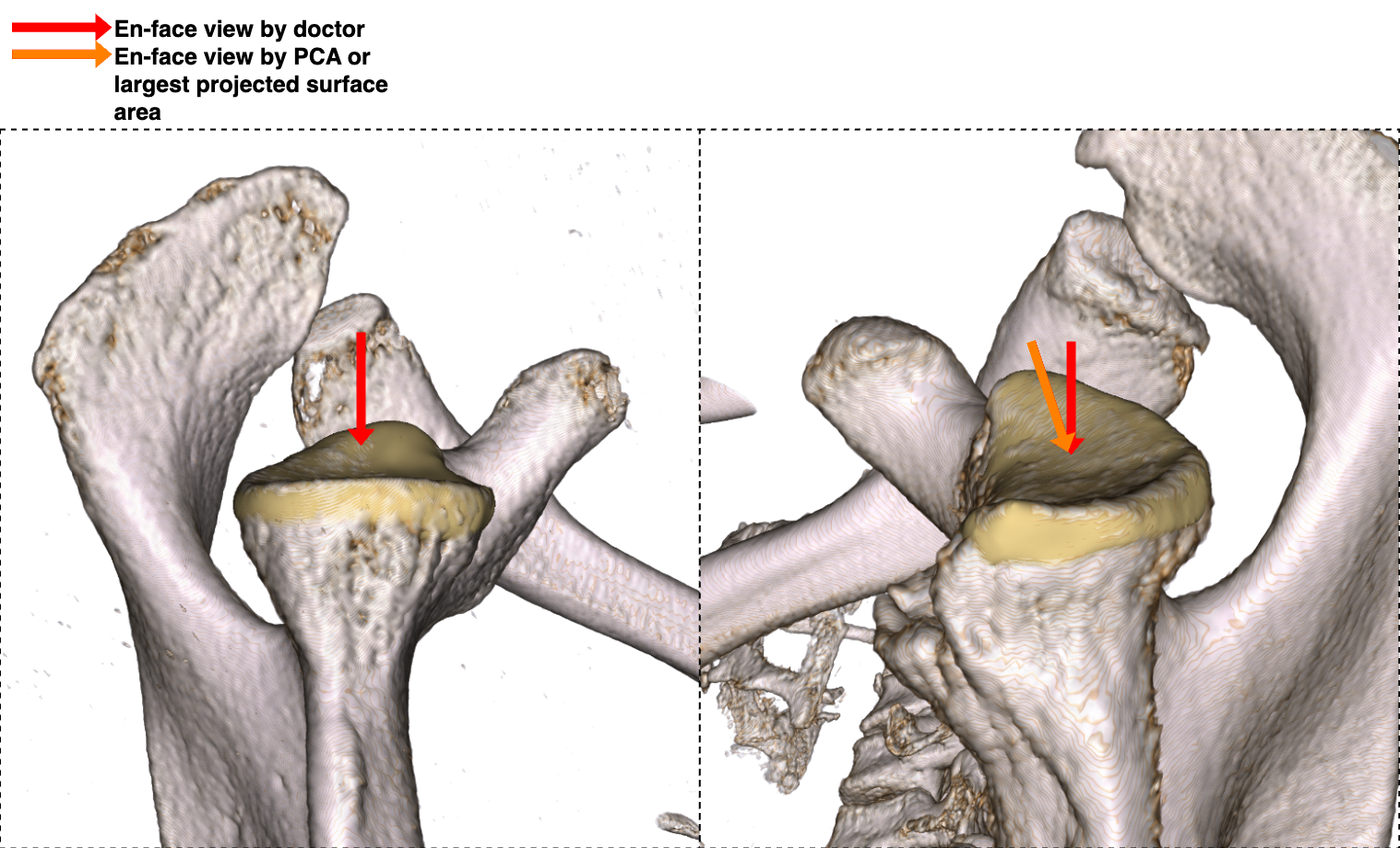}  
    \caption{Comparison between en-face view computed on glenoid with minimal bone loss and high bone loss. Left image shows a healthy intact glenoid bone; Right image shows a glenoid bone with high bone loss. True direction vector selected by doctors are shown in red arrow; direction vector computed by PCA and largest projected surface area is shown in orange arrow.}
    \label{fig:PCA_bad}
\end{figure}

\subsection{Why not PCA / largest projected area?}
Many studies have discussed using PCA or the largest surface area to compute an en-face view to measure glenoid bone loss, either manually or with computer-assisted; however, we found that this approach can be unreliable in cases with severe bone loss (Fig. \ref{fig:PCA_bad}). The cases of high bone loss are the exact scenarios where exact bone loss calculations are needed to appropriate size and plan for bony augmentation. In computer-assisted measurement, this process typically involves segmenting the glenoid bone first, followed by the application of PCA or the largest projected area method. Zhang's study orients the camera along an initial vector and then rotates to find the plane that yields the maximum projected area, thereby finding the true en-face view \cite{Zhang2020TrueEnFaceView}. Haimi's method requires using PCA to find the plane that gives the maximum variance on the glenoid segmentation, and the en-face view is normal to this plane\cite{Haimi2024Automated_2d}. This makes sense because the intrinsic surface of the glenoid bone is generally flat. If you have a good segmentation of the glenoid bone articular surface, it should give you the direction vector identical to the en-face view. However, as shown in Figure \ref{fig:PCA_bad}, in a case of severe bone loss the en-face view computed with PCA is not even close to the true en-face view. Due to intrinsic patterns of bone loss, some bone loss occurs not only on one side at the edge but in the center; In this case, it is not only anteriorly but also from top to bottom in the center. This central (cavitary) glenoid erosion is commonly referred to as a glenoid cup\cite{Kovacevic2014LucentLines}. When a patient experiences bone loss that occurs both on the side and in the center, the articular surface can erode to one side, leading the PCA and the largest projected surface to lean in that direction as well, causing the view of the face to deviate from the expected position. Based on our observation, this type of bone loss isn't uncommon, therefore emphasizing that the computation of en-face view cannot rely on PCA or surface area. Our method utilized topological and anatomical information on the posterior side of the glenoid that is invariant to any type of bone loss, the rim edge points on the posterior side, enabling us to find the same en-face view, regardless of different bone loss types.

\subsection{Rationale for our approach}
Deep learning based methods generally perform the best when trained to specialize in one specific task. Still, we did not optimize the model to output the bone loss, but a segmentation mask and rim points for interpretability in a clinical setting. By training a multi-stage model, one to predict segmentation of glenoid and humerus bone, another to predict rim points where a constraiend circle fitting algorithm fits to for bone loss, it can give surgeons an better grasp at each step as each step imitates exactly what they do in the bone loss measurement guideline: determine what is glenoid and humerus, subtract humerus, determine where is the rim they need to use for circle fitting. With a fully end-to-end deep learning model that maps a CT scan to bone loss regression directly in a black-box framework, not only can we not evaluate each step of the guideline, but there is also less interpretability. We believe our pipeline has a good balance in regressing bone loss and providing interpretability and visualization for the clinical setting. 

One thing to address is RimU-Net's ability to learn the representation of the en-face view and the glenoid rim for circle fitting within a single model. We made this possible because of our data collection method. We first ask the surgeon to reorient the glenoid articular surface to the en-face view, which is normal to the screen. Then, on this plane, surgeons' are required to select the rim points containing the plane's spatial information. Therefore, the models are trained to find rim points already containing the en-face view information. To illustrate this concept, imagine drawing the contour of a bottle in a 3D space from a fixed angle. The paint on that sphere is actually drawn on a 2D space because we are forced to draw on the edge of the object from one angle; therefore, we can utilize the points on that paint to find the en-face direction because the points on the contour should be orthogonal to that angle we were facing.

\end{document}